\documentclass[letterpaper, 10 pt, conference]{ieeeconf}  

\IEEEoverridecommandlockouts                              

\usepackage{stfloats}
\usepackage{float} 
\usepackage[skip=10pt]{caption}

\usepackage{booktabs} 
\usepackage{tabularx} 
\title{\LARGE \bf
Factors Influencing Conversational Engagement in Robot-Delivered Individual Cognitive Stimulation Therapy (iCST) for Dementia in Home Settings}

\author{Emmanuel Akinrintoyo and Nicole Salomons 
\thanks{Emmanuel Akinrintoyo and Nicole Salomons are with  Imperial College London
        {\tt\small ,  e.akinrintoyo23@imperial.ac.uk,
n.salomons@imperial.ac.uk}}%
        }%

\usepackage{graphicx}
\usepackage{dirtytalk}
\usepackage{booktabs} 
\usepackage{amsmath}
\usepackage{graphicx}
\usepackage{pdfpages}
\usepackage{subcaption}
\begin{document}

\maketitle
\thispagestyle{empty}
\pagestyle{empty}

\begin{abstract}
Social robots offer a promising means of supporting cognitive therapies for dementia care by guiding structured conversation and therapeutic activities. However, little is known about the conversational dynamics that emerge during robot-delivered cognitive stimulation therapy (CST) sessions. This study analysed the interaction patterns from robot-delivered individual CST (iCST) sessions conducted with people living with dementia in home settings. Our Co-STAR (Cognitive Stimulation Therapy by an Autonomous Robot) system was deployed in the homes of eight PwDs for one week, who completed 30-minute sessions. Conversational metrics, including words per turn, speech production rate, response duration, response latency, and self-referential language, were analysed to examine how conversational engagement is shaped by prompt personalisation, interaction phase, and participant characteristics. The findings highlight three key interactional properties of robot-delivered iCST. First, personalised prompts significantly increase response duration, self-referential language, and overall engagement compared to generic prompts. Second, conversational behaviour changes within sessions, with a reduction in the verbal output and autobiographical engagement observed during later interaction phases, which suggests cognitive fatigue. Third, first-session conversational metrics were associated with long-term participation, while living situation influenced conversational engagement patterns. These findings provide empirical insights into the factors that shape conversational engagement in robot-delivered iCST. They inform the design of adaptive conversational robots for dementia therapy.
\end{abstract}

\section{INTRODUCTION}


Dementia is a neurocognitive syndrome that manifests as a progressive decline in cognitive and communicative abilities~\cite{taler2008language}. Persons living with dementia (PwDs) experience conversational difficulties. These include word-finding problems, reduced initiation, breakdowns in turn-taking, and greater disfluencies, such as filler words~\cite{bayles1982potential,forbes2005detecting, kempler2005neurocognitive}.

Conversations, such as in therapy sessions, support memory activation, language production, social connectedness and identity expression for PwDs~\cite{spector2003efficacy, kempler2005neurocognitive, taler2008language, conway2000construction, woods2023cognitive}. CST (Cognitive Stimulation Therapy) is an example of such therapies designed for fostering conversational participation with repeated practice
for PwDs~\cite{spector2003efficacy}. CST is one of the leading evidence-based non-pharmacological interventions for individuals living with mild to moderate dementia~\cite{spector2003efficacy, woods2012cognitive}. It has demonstrated benefits for cognition, communication, and quality of life~\cite{spector2003efficacy, woods2012cognitive}. However, delivering CST consistently at scale is resource-intensive. It often relies on trained facilitators and structured group settings.

Recent advances in socially assistive robots (SARs) offer new opportunities to augment or support therapeutic interventions for PwDs~\cite{valenti2015social, akinrintoyo2024co, akinrintoyo2026home}. SARs are being explored as facilitators of cognitive activities, companions, and communication partners. Their potential lies in their ability to deliver therapeutic tasks, scaffold interaction and maintain conversational engagement. Hence, SARs are uniquely positioned to deliver individual CST (iCST)~\cite{yates2015development, orrell2017impact} because they combine embodied presence, conversational interaction, and consistent facilitation. However, limited research has examined the fine-grained conversational dynamics that emerge during robot-delivered therapies.

\begin{figure}[t] 
    \centering
    \includegraphics[width=0.45\textwidth]{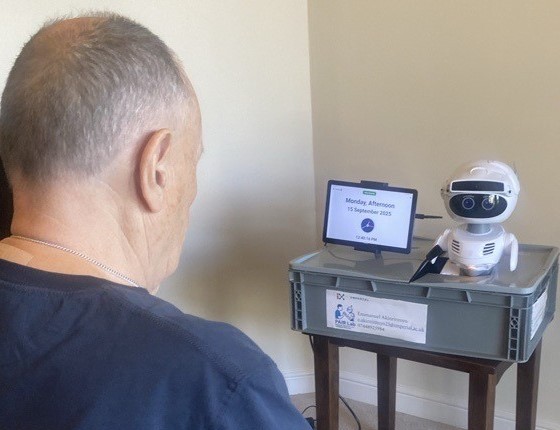}
    \caption{A participant completing an individual cognitive stimulation therapy session in their living room with Co-STAR~\cite{akinrintoyo2026home}.}
    \label{fig:img1}
\end{figure}

This study investigates conversational behaviour in robot-delivered iCST through interaction analysis of audio-recorded therapy sessions with PwDs. It examines how conversational engagement is influenced by prompt personalisation, the interaction phase, and participant characteristics during the interactions. The primary contribution is an empirical characterisation of the factors that shape conversational engagement in robot-delivered therapy and the resulting design implications for adaptive conversational robots.

Furthermore, personalisation is a core component of iCST~\cite{ali2022individual}. The activities are tailored to a person's interests, abilities and life history~\cite{yates2015development, orrell2017impact}. The integration of person-centric prompts can ensure that the tasks are relevant and meaningful for PwDs. Hence, this study investigates whether personalisation can support engagement in robot-delivered therapy. Thus, the following hypothesis was formulated:
\noindent\textbf{H1:} Personalised prompts will elicit higher conversational engagement than generic prompts.


This study also seeks to answer the following research questions: 
\noindent\textbf{RQ1:} Does conversational behaviour change over the course of a therapy session? Analysing within-session dynamics can help to identify changes in participation, responsiveness, or cognitive fatigue during therapy sessions. These patterns could inform phase-based interaction strategies. \noindent\textbf{RQ2:} Can conversational metrics from an initial interaction predict long-term engagement with robot-delivered therapy? Since low adherence is a major hindrance to iCST~\cite{orrell2017impact}, it is vital to understand whether conversational markers from an initial interaction can predict subsequent engagement. This may help to identify PwDs at risk of disengagement. 
\noindent\textbf{RQ3:} Does living situation influence conversational engagement in robot-delivered iCST sessions? Previous iCST research revealed the importance of home and caregiving circumstances in sustaining participation. Since living situations may influence social stimulation, availability of support, and opportunities for interaction, it is essential to understand whether it affects engagement with robot-delivered iCST.

To address the research questions, conversational analysis is performed by analysing interaction metrics, such as response latency, speech production rate, response duration, words per turn, self-referential language, and markers of conversational fluency. The analysis extends beyond outcome-based measures to examine the dynamics of conversational interaction. It includes how participants respond, hesitate and sustain interaction throughout therapy sessions. Therapeutic interactions are dynamic and evolve through conversational sequences~\cite{perakyla2019conversation}. When a robot delivers a therapy session, the structure and flow of conversation may therefore shift in subtle but meaningful ways. It is essential to understand these shifts to inform the design of adaptive robot-delivered therapies that support engagement and participation.


\section{BACKGROUND}
This section reviews conversational challenges in dementia, and social robots supporting PwDs.





\subsection{Conversational Dynamics in Dementia for HRI}
Conversational difficulties are a core feature of dementia~\cite{kempler2005neurocognitive, taler2008language}. They significantly affect everyday conversation as the cognitive impairment progresses. It manifests through difficulties with lexical retrieval, slower processing speed, and coherence difficulties. This leads to longer pauses before responding, increased response latency, and reduced fluency~\cite{kemper2001longitudinal, taler2008language}. 
Notably, these characteristics can also vary significantly between PwDs. 



Repeated exposure to the same conversational partner or interaction format may support adaptation~\cite{kempler2005neurocognitive, taler2008language}. This can include familiarity with the structure, pacing, and expectations of the interaction. Such familiarity can reduce the uncertainty and effort needed to initiate and maintain the conversational responses~\cite{kahneman1973attention, ghafurian2021social}. Hesitations and self-repairs are often associated with lexical search difficulty and processing demands in dementia speech~\cite{almor2009common, davis2009examining}.

Despite these insights, there is limited insight into how observed conversational behaviours connect to sustained effectiveness in robot-delivered therapy. Although prior work has explored linguistic markers of cognitive effort and conversational difficulty in dementia, their potential to predict long-term interaction outcomes has received little attention. Furthermore, social context factors, particularly living situation, remain underexplored; yet understanding their influence on interaction dynamics is crucial for designing adaptive conversational systems for PwDs.

\subsection{Social Robots for Cognitive Stimulation}

SARs are increasingly being explored to support care and interaction ~\cite{feil2005defining, ghafurian2021social} and cognitive stimulation for PwDs~\cite{lu2021effectiveness}. For instance, companion robots such as PARO have shown emotional comfort and interaction in residential care settings~\cite{vsabanovic2013paro}. Other conversational robots have facilitated reminiscence, guided activities, or provided structured prompts during cognitive stimulation tasks~\cite{lee2020four, cruz2020social}. The systems seek to reduce caregivers' burdens while providing consistent interaction.

Previous research has focused on accessibility, acceptance, usability, and emotional response rather than examining the conversational dynamics between PwDs and SARs~\cite{wu2016attitudes}. Less is known about how conversational features such as prompt type, response timing, and words per turn influence interactions in robot-delivered cognitive stimulation sessions. Understanding these interaction patterns can aid the design of systems that support sustained and meaningful conversation. 
These conversational dynamics that emerge during robot-delivered therapy are investigated in this work through in-home iCST sessions delivered for one week. 

\section{METHODOLOGY}

In this section, we describe the robotic iCST system, iCST activities, session design, and participants' details.

\subsection{System}
An autonomous robotics system was designed following multiple design consultations and evaluations with dementia stakeholders~\cite{akinrintoyo2025home}. This included PwDs, formal and informal caregivers, dementia well-being managers, occupational therapists, dementia nurses and mental health practitioners, as reported in our previous work~\cite{akinrintoyo2025home}. Co-STAR (Cognitive Stimulation Therapy by an Autonomous Robot)~\cite{akinrintoyo2026home}, the final system was then deployed for evaluation in participants' homes. 

Co-STAR (Fig. \ref{fig:img1}) consisted of a Misty robot. It included a tablet that displayed relevant images and text spoken by the robot. Users could initiate a session by clicking on the \say{Start} button on the tablet. The system also included a microphone and a mini-computer. The mini-computer was used for all the processing tasks. It helped to deploy a private cloud architecture, such that all data processing and storage were performed locally at home. No data was processed on the cloud or by third-party service providers. It was also aided by using a local speech-to-text system. The audio recordings of the conversations were stored locally on the mini-computer. No video data was collected. All audio spoken by Co-STAR was generated before deployment.

\subsection{iCST Activities}

Five iCST activities were developed for in-home sessions. The sessions provide an opportunity for PwDs to engage in personalised cognitive stimulation that is tailored to their interests. Word Association involved recognising a common theme between multiple images. The Popular Places activity helped recollect famous landmarks they might have seen or visited. Object Categorisation required the grouping of similar images, such as everyday objects. Famous Faces involved recognising celebrities and prominent figures. Common Sayings involved completing familiar proverbs. The activities aimed to promote reminiscence, communication, reinforce word retrieval, and support ongoing cognitive activation.

Each structured activity was designed to be turn-based and conversational for the PwD. Co-STAR facilitated the therapeutic activity through spoken prompts that included open-ended questions and reminiscence-based discussions. While the prompts encouraged verbal responses, PwDs could respond at their own pace. Follow-up prompts were provided for each question to sustain the conversation. Personalised prompts were integrated into the sessions with questions referencing the user's past experiences, hobbies and memories.  
  
\subsection{iCST Session}
Each iCST session was designed to last $30$ minutes. A session commenced with a greeting and introduction. Co-STAR provided the user with orientation information, such as day, date and time. A detailed description of the activity is then provided to the user. This included what to expect during the session and how to participate. The user is also reassured that it is neither a quiz nor a test. The themed activity then commenced with Co-STAR guiding the participant through the session. It concluded by thanking the user. An adaptive response latency mechanism was integrated into the robot. This allowed it to learn and adjust its response timing and turn-taking behaviour based on the user’s speaking pace observed during the session.

\begin{table}[t]
  \caption{List of participants: pseudonym, age, gender, living status, and sessions completed.}
  \label{tab:participants}
  \centering
  \small
  \begin{tabular}{@{}l c c l c c@{}} 
    \toprule
    \textbf{Pseudonym} & \textbf{Age} & \textbf{Gender} &  \textbf{Living} & \textbf{Sessions} \\
    \midrule
    Ray        & 83 & M    & Son     & 5 \\
    Julia      & 77 & F    & Partner & 3 \\
    Georgia    & 78 & F    & Alone   & 1 \\
    Esther     & 81 & F    & Alone   & 5 \\
    Benjamin   & 59 & M    & Partner & 1 \\
    Gideon     & 78 & M   & Partner & 8 \\
    Priscilla  & 94 & F   & Alone   & 5 \\
    Martha     & 69 & F   & Alone   & 3 \\
    \bottomrule
  \end{tabular}
\end{table}

\subsection{Participants}
Ethical approval was obtained for this study (ethics no: 7091501). Eight people with a clinical diagnosis of dementia were recruited through the expert guidance of dementia care managers of dementia memory clinics and independent living centres. The recruitment process for selection was based on the guidance of the care managers. Each person living with dementia had a sufficient level of capacity to provide informed consent. This was supported by consent from their caregivers (spouse or partner), where applicable. The profiles of the participants are presented in Table \ref{tab:participants} with pseudonyms to replace their real names. They had an average age of 79.67 years. Each participant was compensated with $50$ pounds.
 

Co-STAR~\cite{akinrintoyo2026home} was deployed in the homes of PwDs to deliver iCST for a week per PwD. Participants were provided with an information sheet that described the study details, what to expect and who to contact for any queries. Participants were provided with a guide for using the system.

\section{Data Analysis and Metrics}
This section describes the computational pipeline and metrics used to analyse the audio recordings of PwDs during the week-long robot-delivered iCST deployment. 
Conversational behaviour analysis in dementia requires metrics that capture multiple dimensions of interaction. Measures such as response latency, words per turn, and self-corrections can provide insights into how a PwD engages with conversational prompts~\cite{kemper2001longitudinal, taler2008language}. They can reveal how cognitive demands influence speech production rate and conversational adaptation patterns. Robot-delivered therapy settings should understand these interaction dynamics for adaptive designs that support long-term use.

\subsection{Audio Processing and Transcription} 
Thirty-one therapy sessions were conducted in total (Table \ref{tab:participants}). Four recordings (Ray: $1$, Julia: $2$, Gideon: $1$) were unavailable due to technical issues during data capture and were therefore excluded. The analysis was conducted on the remaining $27$ sessions. A multi-state speech analysis pipeline was used to produce high-fidelity transcripts and accurately segment conversational turns. Speech recognition was performed using Whisperx~\cite{bain2023whisperx} and WhisperD~\cite{akinrintoyo2025whisperd}. WhisperX is an advanced automatic speech recognition (ASR) system that is optimised for fast and accurate transcription. It converted audio to text with phoneme alignment. This re-aligned the transcripts at the individual-word level to ensure precise start and end timestamps. WhisperD is the state-of-the-art speech recognition system for dementia speech. It includes filler words such as \textit{\say{uh}} and \textit{\say{um}} in its output transcriptions, which are prevalent in dementia speech. This preserves the natural conversational characteristics and patterns with diagnostic relevance. The transcripts were manually verified and sanitised to remove any personally identifiable data.

Speaker diarisation was performed using a Hugging Face pipeline~\cite{Plaquet23, Bredin23}. It identified distinct voices and assigned a speaker label (Participant or Robot) to each turn. This captured the turn-taking dynamics in the audio segments. Using these timestamps and speaker labels, a turn-by-turn conversational map was constructed for each interaction session. Co-STAR's prompts were provided as ground truth.

Session interaction metrics were extracted to characterise the overall interaction structure. The metrics include the total number of conversational turns (robot, participant), the total number of words spoken by the participant and the total participant speaking duration. These measures capture the overall volume and structure of speech produced during the iCST sessions. A locally deployed LLama-3 model was leveraged to analyse each conversational turn. Intent classification was performed to identify conversational signals and cognitive markers, such as cognitive breakdown, hesitation and self-reference words. The failure diagnostics were examined for every breakdown or long silence. Each participant's response was processed independently using structured prompts that guided the model to identify the markers based on predefined definitions. The LLama-3 model was also used to categorise Co-STAR's prompts as personalised or generic. All the LLM outputs were manually verified for consistency.

\subsection{Speech Production Metrics}
Speech production metrics capture how speech is produced during interactions. They include:

\noindent\textbf{Words per Turn (WPT):} Word count quantifies the number of words produced by a participant in a single response turn after a prompt by Co-STAR. 

\noindent\textbf{Response Duration:} This is the total time spent speaking in a single response. 

\noindent\textbf{Speech Production Rate (SPR):} It was operationalised as words per second to provide a temporal measure of conversational fluency. 

\noindent\textbf{Response Latency:} This is the temporal gap between when the robot's prompt ended and when the user started speaking. This may reflect the user's cognitive processing time.



\subsection{Fluency and Repair Markers}
Fluency and repair markers capture disruptions in the flow of conversation and the strategies used to correct difficulties in speech production rate. They include: 

\noindent\textbf{Hesitation:} Disfluencies such as filler words (\textit{\say{uh}} and \textit{\say{um}}) and hedging language (\textit{\say{maybe}} and \textit{\say{not sure}}) were analysed as indicators of conversational hesitation and uncertainty. These expressions are commonly associated with speech planning difficulty and level of certainty in speech~\cite{levelt1993speaking, davis2009examining}.

\noindent\textbf{Self-Correction:} Self-corrections represent successful repair attempts during speech production rate. Repair phrases such as \textit{\say{I mean}}, \textit{\say{no wait}}, \textit{\say{excuse me}}, \textit{\say{actually}}, \textit{\say{sorry}} were used to detect instances where participants attempted to reformulate their speech~\cite{levelt1993speaking, davis2009examining}. 

\noindent\textbf{Cognitive Breakdown:} Cognitive breakdowns indicate retrieval difficulty or disrupted speech without a successful repair. The occurrence of specific phrases such as \textit{\say{I don't know}}, \textit{\say{I can't recall}}, \textit{\say{I forgot}}, and \textit{\say{I don't/can't remember}} was used to identify instances of retrieval difficulty. These expressions may reflect lexical access problems in dementia~\cite{almor2009common}.

\subsection{Conversational Engagement Markers}
Conversational engagement markers indicate how actively and meaningfully a person participates in a conversation:

\noindent\textbf{Self-Reference Rate:} Self-reference measures the proportion of responses that contain first-person pronouns (e.g., “I”, “my”, “we”)  or autobiographical markers.


\noindent\textbf{Positive Sentiment:} This measures the proportion of the speaker's responses that contain positive emotional expressions during an interaction.



\subsection{Statistical Analysis}
Non-parametric statistical tests were used throughout the analysis. This is due to the cohort size and distribution of the conversational metrics. The Mann–Whitney $U$ test was used to compare continuous variables (e.g., latency, speech production rate, duration). It compared the rank ordering of the observations rather than their raw values. It is robust to outliers and non-normal distributions. Chi-squared tests were used to compare proportions between groups. The Kruskal-Wallis $H$ test was used to assess differences across the entire cohort. It is the non-parametric equivalent of an ANOVA. 



\begin{figure*}[t]
\centering
\includegraphics[width=\linewidth]{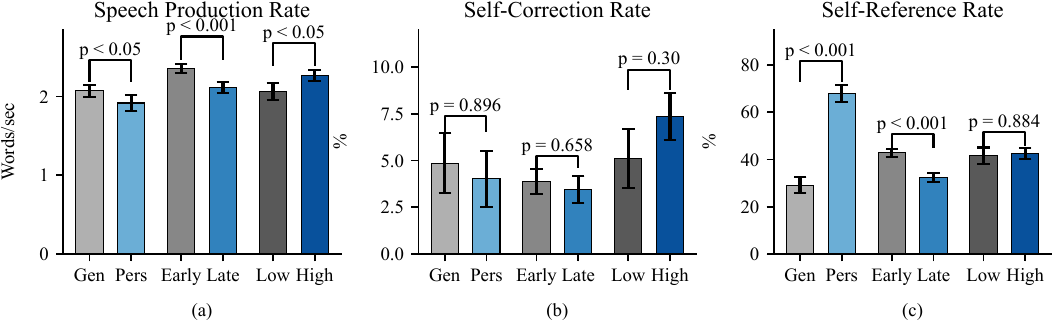}
\caption{Effects of conversational dynamics on participant output: (a) Speech production rate, (b) Self-correction rate, and (c) Self-reference rate. Comparisons include generic (Gen) versus personalised (Pers) prompts, early versus late session phases, and low versus high engagement cohorts. Error bars represent $\pm$ standard error.}


\label{fig:plots}
\end{figure*}



\section{Results and Discussion}
Across the deployment phase, participants engaged in robot-delivered therapy sessions with an average interaction time of 32 minutes and 42 seconds. 

\subsection{Effects of Personalised vs Generic Prompts}
Co-STAR's prompts were categorised as \textit{personalised} or \textit{generic}. Responses to each prompt type were compared to investigate how personalisation influenced conversational participation. The comparison was restricted to open-ended questions. Yes/no answers were excluded. The personalised prompts included those linked to a participant’s life, preferences, or experiences. They included questions such as \textit{\say{What did you use to do for work?}} and \textit{\say{Do you enjoy having plants at home?}}. Generic prompts consisted of neutral or task-based questions. They included questions such as \textit{\say{What do you think curtains and lamps have in common?}} and \textit{\say{Which two or three objects do you think are similar?}}.

\subsubsection{Results}
The number of turns was relatively balanced across conditions ($174$ vs $185$). There was an average of $24.29$ WPT for personalised compared with $17.32$ WPT for the generic case ($p = 0.113$). The largest difference was observed in the response duration (Fig. \ref{fig:plots}(a)). On average, participants talked for longer when responding to a personalised prompt compared to generic prompts ($17.3$ vs $10.07$ seconds, $p < 0.01$). This represents a $70\%$ difference in speaking time. In addition, speech production rate differed between the two prompt types. Participants had a lower speech production rate (Fig. \ref{fig:plots}(b)) for personalised prompts than generic prompts ($1.92$ vs $2.08$ words per second, $p < 0.05$). There was no statistically significant difference in response latency between the two prompt types ($1.91$ vs $1.70$ seconds, $p = 0.167$).

Non-statistically significant differences were recorded in the hesitation rates of personalised prompts ($8.6\%$ vs $3.2\%$, $p = 0.571$), self-correction ($4.0\%$ vs $4.9\%$, $p = 0.896$) and cognitive breakdown ($5.7\%$ vs $3.2\%$, $p = 0.372$), compared with generic prompts. 

Personalised prompts produced significantly higher levels of autobiographical framing. The self-referential language percentage (Fig. \ref{fig:plots}(c)) significantly increased for personalised prompts ($67.8\%$ vs $29.2\%$, $p < 0.001$) compared to generic prompts (Fig. \ref{fig:plots}(c)). This increase was also associated with a higher positive sentiment ($29.3\%$ vs $13.5\%$, $p < 0.001$) for personalised prompts than observed in generic prompts. 




\subsubsection{Discussion}
The comparison revealed several consistent patterns that indicate that personalisation influences conversational behaviour. Participants spent more time responding to personalised prompts, used significantly more self-referential language, and expressed more positive sentiment in their sentences. Therefore, we found evidence to support \textbf{H1:} \textit{Personalised prompts will elicit higher conversational engagement than generic prompts.}

Response duration had the most notable increase. Participants spoke for approximately 70\% longer for personalised prompts. Notably, this occurred without a corresponding increase in latency. This suggests that the participants were providing elaborate responses once they became engaged rather than taking longer to start speaking. The higher number of words per turn also supports the interpretation that personalisation fostered greater conversational contribution among participants. 

Personalised prompts also encouraged greater use of self-referential language and more positive sentiment. This may suggest that participants drew more heavily on their own experiences and perspectives when responding. While speech production was slower with personalised prompts, the absence of significant differences in hesitation rates, self-corrections, and cognitive breakdowns may indicate that the slower pace reflected increased reflection and autobiographical recall rather than increased difficulty. Taken together, these findings suggest that personalisation increases participant involvement. It also promotes a more personal conversational style with longer and more positive responses.



\noindent\textbf{Design Implication:} 
Robots should consider prioritising personalised prompts over generic prompts. The prompts should connect to PwDs' autobiographical experiences.

\subsection{Early vs Late Interaction Phase}
Given a 30-minute session length for the iCST intervention, the two halves of the session were compared to assess differences in within-session conversational dynamics. The first $15$ minutes represented the early interaction phase, while the final 15 minutes represented the later phase of the session. This segmentation ensured sufficient conversational turns for a reliable comparison while enabling the observation of temporal changes in conversational participation. The comparison was done to investigate how conversational behaviour changes during a therapy session. 

\subsubsection{Results}
During the first $15$ minutes, participants showed relatively high levels of proactive participation as reflected in higher WPT ($18.83$ vs $14.74$, $p < 0.001$). This represents a $21\%$ decrease in words produced per turn. Non-statistically significant results were obtained for response duration ($14.25$ vs $11.15$ seconds, $p = 0.361$). Speech production rate also declined during the session ($2.36$ vs $2.12$ words per second, $p < 0.001$). No statistically significant difference was observed in the response latency ($1.899$ vs $2.241$ seconds, $p = 0.223$). There were no statistical differences for any of the fluency and repair markers, including self-correction rate ($3.9\%$ vs $3.4\%$, $p = 0.658$). Conversational engagement markers indicated a reduction in autobiographical engagement.  This engagement was indicated by a decline in self-reference rate ($42.8\%$ to $32.5\%$, $p < 0.001$). 


\subsubsection{Discussion} 
Participants exhibited a higher speech production rate and produced longer responses during the first phase of the interaction. This indicates more active participation at the start. The conversational narratives often included dependent clauses and temporal markers (e.g., \textit{\say{One of the dogs I had was an amazing dog, he knew exactly where I was when I was and when he wanted to come out...}} - Gideon). Responses became sparse, direct, and reliant on procedural memory in the later phase (e.g., \textit{\say{Yes, it was good}}, \textit{\say{I liked that}}). The within-session analysis may be consistent with reduced engagement or cognitive fatigue in the final phase. While decreases in speech production rate, self-reference rate and WPT were observed, participants appeared to produce shorter and less autobiographical responses in the later phase rather than sustained elaboration. These findings address \textbf{RQ1}: \textit{Does conversational behaviour change over the course of a therapy session?} The results suggest that conversational pace and utterance length decline over time. 


\noindent\textbf{Design Implication:} Robot-delivered iCST may benefit from phase-based interaction strategies. Early interaction periods may support higher-cognitive-load tasks, such as autobiographical recall or exploratory conversation, which are core components of CST~\cite{woods2012cognitive}. Later phases may benefit from lower cognitive load, such as familiar topics or simpler conversational structures. 
Moreover, shorter but more frequent sessions may be more effective.

\subsection{First-Session Engagement Indicators}
An analysis was conducted to examine whether the long-term engagement with the robot-delivered intervention can be predicted at early interaction. Due to the uneven distribution of interaction frequency across the eight participants, the participants were split into two groups: (1) low-engagement participants who completed three or fewer sessions ($N = 4$), and (2) high-engagement participants who completed more than three sessions ($N = 4$). 
First-session conversational metrics were compared between the groups to answer \textbf{RQ2}. This analysis examines whether initial conversational behaviour is associated with long-term engagement. 


\subsubsection{Results}

No statistically significant differences were observed in the WPT ($23.90$ vs $17.76$, $p = 0.321$) or response duration ($18.19$ vs $9.80$, $p = 0.060$). However, low-engagement participants had a lower speech production rate ($2.066$ vs $2.270$ words per second, $p < 0.05$). A notable difference was observed in the response latency. Low-engagement participants also required more time to begin speaking. This was indicated by their higher mean response latency ($2.70$ vs $1.63$ seconds, $p < 0.001$). 

Fluency and repair markers also differentiated the groups. Low-engagement participants had higher hesitation rates ($8.2\%$ vs $3.7\%$, $p < 0.05$). However, no statistically significant differences were observed in self-corrections ($5.1\%$ vs $7.4\%$, $p = 0.30$) and cognitive breakdown ($4.6\%$ vs $5.7\%$, $p = 0.561$). There were no statistically significant differences in the self-reference rates ($41.5\%$ vs $42.5\%$, $p = 0.884$). Low-engagement participants showed higher positive sentiment ($21.5\%$ vs $14.7\%$, $p < 0.05$). 


\subsubsection{Discussion}
The analyses revealed several differences in conversational behaviours between the two participant groups. First-session behaviour analysis revealed a distinct conversational profile for participants who did not continue with subsequent sessions. They exhibited longer response latencies, slower speech production rates and higher hesitation rates. The findings suggest that first-session conversational behaviour may provide early indicators of long-term engagement. This offers preliminary evidence in support of \textbf{RQ2}: \textit{Can conversational metrics from an initial interaction predict long-term engagement with robot-delivered therapy?}

A possible explanation is that early conversational metrics may be associated with interaction difficulty levels. Previous research has shown that hesitation frequency increases under higher cognitive load. This suggests that longer response latencies, slower speech production rates, and higher hesitation rates may reflect greater cognitive effort during interaction~\cite{betz2023cognitive}. Notably, low-engagement participants exhibited higher positive sentiment. This suggests that affective expression and sustained engagement may not directly align. Long-term engagement was not associated with differences in self-referential language use unlike personalisation and session phase. 

\noindent\textbf{Design Implication:} If a conversational robot detects signs of increased response latency, it could adapt its interaction strategy in real time. For example, the robot could simplify its prompts or increase the contextual cues. Such adaptive mechanisms may help reduce cognitive load during early interactions while supporting long-term engagement with robot-delivered therapy.

\subsection{Influence of Living Situation on Engagement}
To address \textbf{RQ3}, participants were categorised based on their living conditions (Table \ref{tab:participants}). Among the eight participants, four lived alone, and four lived with a family caregiver. These were compared to examine whether their social living situation influenced conversational engagement. 

\subsubsection{Results}
Participants living alone had significantly longer responses ($21.9$ vs $15.5$ $WPT$, $p < 0.001$) and a longer response duration ($16.8$ vs $9.1$ seconds per turn, $p < 0.001$). In addition, they had a lower speech production rate ($2.16$ vs $2.47$ words per second, $p < 0.001$) and response latency ($1.55$ vs $2.24$ seconds, $p < 0.001$).

Non-statistically significant differences were observed for hesitation ($5.59\%$ vs $4.43\%$, $p = 0.245$). Other speech patterns included increased self-corrections rates ($7.6\%$ vs $2.3\%$, $p < 0.001$) and cognitive breakdown ($8.8\%$ vs $3.2\%$, $p < 0.001$) for participants living alone. 

Lastly, participants living alone utilised more self-referential language ($47.9\%$ vs $32.2\%$, $p < 0.001$). Non-statistically significant differences were observed for positive Sentiment ($20.32\%$ vs $17.39\%$, $p = 0.101$).

\subsubsection{Discussion}

The findings suggest an association between living situation and conversational engagement during robot-delivered therapy. Participants living alone gave longer responses, spent more time speaking and used more self-referential language than those living with a family caregiver. This offers a nuanced answer to \textbf{RQ3}: \textit{Does living situation influence conversational engagement in robot-delivered iCST sessions?} Thus, robot-delivered interventions interactions may provide opportunities for autobiographical expression.


The positive patterns observed in participants living alone were accompanied by higher cognitive breakdowns and self-corrections. This may indicate greater cognitive effort and retrieval difficulty. It suggests a trade-off between increased narrative engagement and greater conversational difficulty among participants living alone. Conversely, participants living with a family caregiver provided shorter responses and spoke faster. One possible explanation is that they may have an increased familiarity with regular social interaction. It could also be due to the interruptions from family caregivers that were present during the sessions.

\noindent\textbf{Design Implication:} These findings highlight the importance of considering the user's social context when designing social robots for therapy settings. Persons living alone may benefit from robots that encourage extended storytelling and autobiographical recall. They may have fewer opportunities for daily social interaction compared to those who live with others. Participants living with others can utilise robots as conversational facilitators instead. Overall, the participants' living conditions should be part of their user profiling in the system design. This can aid the development of adaptive conversational behaviours in robots that respond to users’ social environments for better engagement and therapy outcomes.

\section{Conclusion}
This study examined conversational behaviour in robot-delivered iCST for PwDs. The findings demonstrate that personalisation, the interaction phase, and participant characteristics shape conversational engagement. While the findings are limited by the small sample size (eight participants), they identify promising directions for robot-delivered therapy. Future work will involve longer in-home deployments with larger cohorts to validate these findings.

\bibliographystyle{IEEEtran}
\bibliography{IEEEabrv,bibliography}

\end{document}